\documentclass[conference]{IEEEtran}
\IEEEoverridecommandlockouts

\usepackage{cite}
\usepackage{amsmath,amssymb,amsfonts}
\usepackage{graphicx}
\usepackage{textcomp}
\usepackage{xcolor}
\usepackage{fontawesome5}

\usepackage{hyperref}

\usepackage{algorithm}

\begin{document}

\title{Learning Task and Motion Plans from Real Demonstrations with Hybrid Flow Matching}

\author{Zuleika Redondo Garcia$^{*,1}$, Andreu Matoses Gimenez$^{*,1}$, and Javier Alonso-Mora$^{1}$%
\thanks{$^{*}$Equal contribution.}%
\thanks{This work is supported by the European Union through the ERC project INTERACT (Grant agreement No. 101041863). Views and opinions expressed are however those of the author(s) only and do not necessarily reflect those of the European Union or the European Research Council Executive Agency. Neither the European Union nor the granting authority can be held responsible for them.}%

\thanks{$^{1}$The authors are with the Department of Cognitive Robotics,
        Delft University of Technology, 2628 CD Delft, The Netherlands.
        {\tt\small \{A.MatosesGimenez, J.AlonsoMora\}@tudelft.nl}}%
\thanks{This work has been submitted to the IEEE for possible publication. Copyright may be transferred without notice, after which this version may no longer be accessible.}%
}

\maketitle

\begin{abstract}
Long-horizon mobile manipulation requires a task plan and the motion that executes it. Generative planners trained on demonstration produce both in one pass, requiring neither a symbolic domain nor search. To date, however, they have relied on thousands of scripted demonstrations of fixed-base arms and executed open loop.
This paper presents a hybrid flow matching planner: a single network generates the symbolic plan with masked discrete flow matching and the motion trajectory with continuous flow matching. Unlike prior generative planners, we aim to learn from a much smaller set of demonstrations and to execute the plan in closed loop. Two properties of the data compensate for the small dataset. 
A demonstration resumed from any of its intermediate actions is itself a demonstration, which multiplies the training samples and enables replanning after every action. Objects of the same kind are interchangeable, which turns demonstrations of one goal into demonstrations of every permuted goal. Our base implementation produces valid plans on 68\% of held-out scenes; a training and generation scheme for the discrete plan raises this to 76\%, and replanning after every action raises the task completion rate from 40\% to 53\% in a kinematic simulation.
The planner matches the task completion rate of motion-only flow matching policies while additionally providing the symbolic plan, and it outperforms previous hybrid diffusion formulations on both task completion and plan validity. We validate the planner on a real mobile manipulator. 
Videos and project page:
\faGlobe~{\urlstyle{same}\href{https://andreumatoses.github.io/research/hybrid-flow-planning}{\nolinkurl{andreumatoses.github.io/research/hybrid-flow-planning}}}

\end{abstract}


\section{Introduction}
\label{sec:intro}

\begin{figure}[t]
    \centering
    \includegraphics[width=\linewidth]{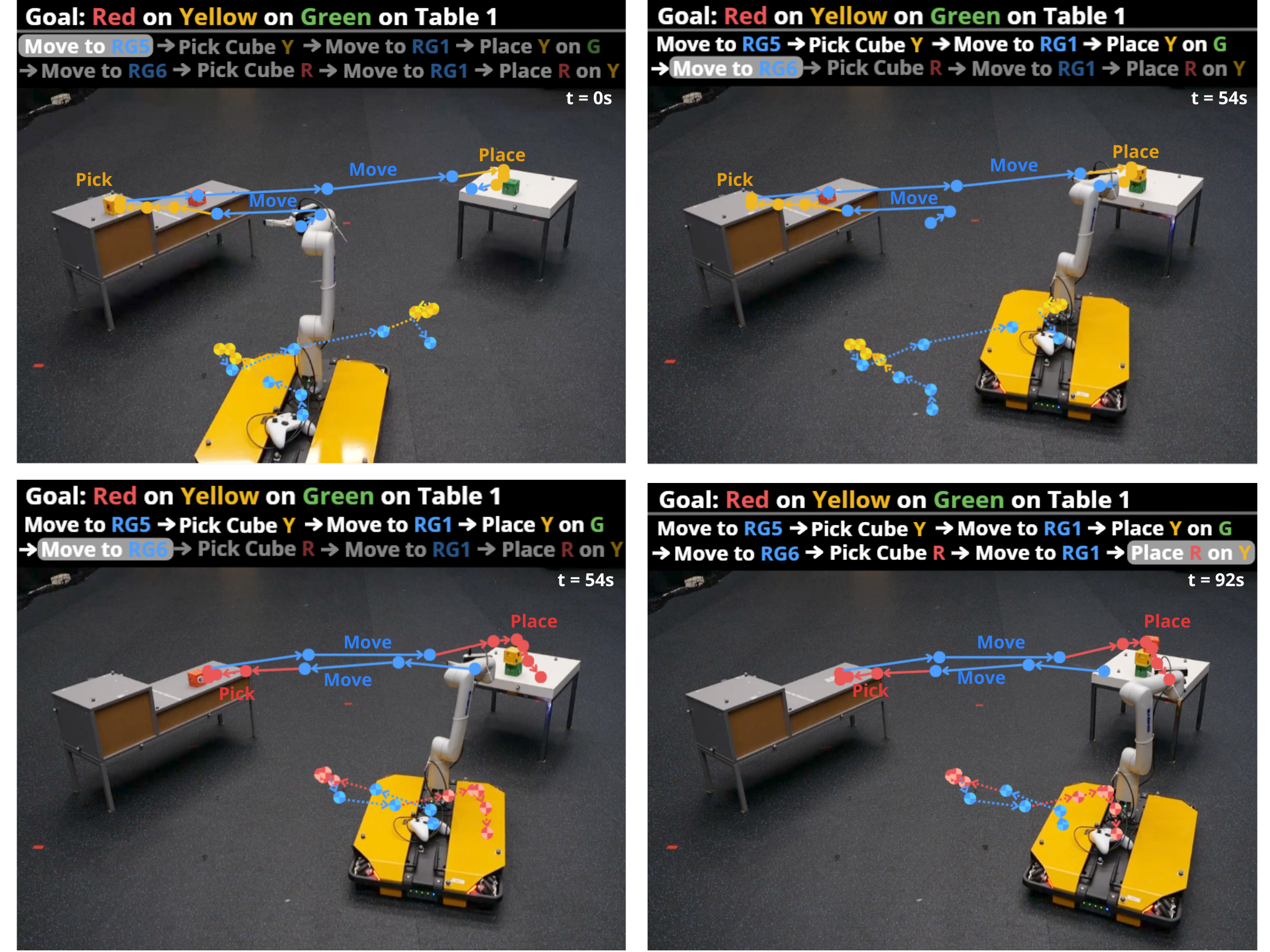}
    \caption{The planner on the real robot building the tower R on Y on G. Four frames of one run: the start and end of the first half of the task (top) and of the second half (bottom). Each frame shows the plan the planner generated from that scene, with the action in progress highlighted, and the generated waypoints of the base and end effector. The planner generates the waypoints of the whole task; for readability only those of the half shown are drawn.}
    \label{fig:teaser}
\end{figure}

Mobile manipulation tasks such as tidying a room or assembling a structure require a robot to decide what to do and how to move. To stack three cubes on a target table (Fig.~\ref{fig:teaser}), a robot has to choose which cube to pick first, from which side of which table, in which position to place it. It also has to produce the base and arm motion for each choice. The two levels depend on each other: a cube can only be reached from certain sides of a table, and a covered cube cannot be picked. Task and motion planning (TAMP) produces both the action sequence and the motion that realizes it while keeping them consistent \cite{garrett2021integrated}.

Classical TAMP methods search over symbolic action sequences and call a motion planner to check each candidate \cite{kaelbling2011hierarchical,srivastava2014combined,toussaint2015logic,garrett2020pddlstream}. They need a hand-written model of the task: predicates, action operators, and samplers for continuous parameters such as grasps and base placements. The search is also expensive, which limits replanning when the scene changes. Learning can guide the search \cite{chitnis2016guided,driess2020deep} or distill planner solutions into policies \cite{dalal2023imitating}, but both still depend on the planner and its model.

Generative models trained on demonstrations take a different route. The task knowledge enters through the demonstrations, so no planning domain is written by hand. A plan then costs one generation pass with a fixed number of network evaluations instead of a search over candidate plans. Diffusion and flow matching policies capture the multimodal distribution of demonstrated motion and generate whole trajectories \cite{janner2022planningdiffusionflexiblebehavior,chi2024diffusionpolicyvisuomotorpolicy,flow-matching-lipman,black2024pi0}. On long tasks with several distinct decisions, however, the model blends different behaviors and the trajectory commits to none of them. H{\o}eg et al.~\cite{hoeg_hybrid_2025} address this with a discrete and a continuous diffusion process in one network that generate a symbolic plan together with the trajectory, and show that the joint generation outperforms trajectory-only generation.

The space of task and motion plans is combinatorial, as each possible action sequence must be grounded in the geometry of the scene. Prior joint symbolic and continuous generative planners have therefore relied on hundreds to thousands of demonstrations per task, often generated by scripted experts on fixed-base manipulators and executed open loop \cite{hoeg_hybrid_2025}. 
For real mobile manipulators, however, demonstrations are typically collected through teleoperation and cannot be generated at that scale. This limited coverage raises three challenges: 
generalization to unseen scenes and plans, generation of variable-length symbolic plans, and error accumulation during open-loop execution.

We address these challenges with a hybrid flow matching planner for mobile manipulation trained on about one hundred real demonstrations. Given an observation of the scene, a single network simultaneously generates a discrete symbolic plan and the continuous trajectories of both the mobile base and the end effector. 
To improve data efficiency, we augment each demonstration with its suffixes, the remaining plan and trajectory after each of its actions, which also enables the planner to replan after every action. We also exploit object interchangeability through object permutations, a consistent renaming of the objects in plan and observation, to obtain goal-conditioned demonstrations.
We further constrain the generated plans to well-formed action sequences, adapt training to variable-length plans, and automatically balance the discrete and continuous losses.

We evaluate the method on a mobile manipulator building a three-cube tower from 109 teleoperated demonstrations. Generated plans are validated with a symbolic model, executed in kinematic simulation, and evaluated on the real robot. We compare against motion-only flow matching and the hybrid diffusion planner of H{\o}eg et al.~\cite{hoeg_hybrid_2025}, and ablate the proposed components.

The contributions of this paper are:
\begin{itemize}
    \item A hybrid flow matching planner that jointly generates the symbolic plan and the motion of a mobile manipulator, with a training and generation scheme that improves symbolic plan validity and task completion.
    \item Two augmentations that exploit the structure of task and motion demonstrations, which allow replanning after every action and goal conditioning without new demonstrations (Section~\ref{sec:augmentation}).
    \item An evaluation on real demonstrations against motion-only flow matching and hybrid diffusion~\cite{hoeg_hybrid_2025}, with ablations and a deployment on the real robot (Section~\ref{sec:results}). %
\end{itemize}

\section{Related Work}
\label{sec:related_work}

\paragraph*{Task and motion planning}
Task and motion planning (TAMP) integrates symbolic task planning with geometric reasoning to ensure that high-level action sequences are physically feasible \cite{garrett2021integrated}. Existing approaches couple these levels in different ways. Hierarchical methods progressively interleave task and motion planning \cite{kaelbling2011hierarchical}, while planner-independent frameworks allow symbolic planners to invoke black-box geometric samplers \cite{srivastava2014combined,garrett2020pddlstream,gimenez2026cross}. Logic-geometric programming instead optimizes the action sequence and its associated motion jointly \cite{toussaint2015logic}. Learning has also been used to guide the planning process, for example by ranking candidate refinements \cite{chitnis2016guided} or predicting promising action sequences before geometric verification \cite{driess2020deep}. Other work distills planner-generated solutions into visuomotor policies \cite{dalal2023imitating}. These approaches generally retain an explicit planning model, either during execution or when generating training data. In contrast, we consider learning symbolic plans and their associated motions directly from teleoperated demonstrations.

\paragraph*{Generative models for robot motion and planning}
Diffusion models generate multimodal trajectories from demonstrations, for planning \cite{janner2022planningdiffusionflexiblebehavior,carvalho2023motion} and for visuomotor control \cite{chi2024diffusionpolicyvisuomotorpolicy}. Flow matching trains the same class of models with a regression objective and supports sampling in only a few integration steps \cite{flow-matching-lipman}; it has since been adopted for robot policies \cite{braun2024riemannian,black2024pi0}. For long-horizon tasks, however, task structure is often provided separately. Generative skill chaining composes diffusion models according to a given skill sequence \cite{mishra2023generativeskillchaininglonghorizon}, SkillDiffuser learns discrete skill abstractions from language instructions \cite{Liang_2024_CVPR}, and LTLDoG constrains sampling with temporal logic \cite{dog}. Inside classical TAMP, diffusion models have served as samplers for continuous parameters \cite{yang2023compositional,fang2024dimsam}, while language models can provide high-level plans that are subsequently grounded by a diffusion model \cite{sharan2024plan}. In these approaches, the symbolic task structure is provided, inferred, or planned separately from the continuous motion. Our planner instead generates both jointly.

\paragraph*{Joint discrete and continuous generation}
Discrete diffusion and flow models generate token sequences by progressively unmasking an initially masked sequence \cite{austin2021structured,chang2022maskgit,sahoo2024simple,gat2024discrete}. Campbell et al.~\cite{campbell2024generative} combine such a masked discrete flow with a continuous flow to co-design protein sequences and structures, while Diffusion Forcing assigns individual noise levels to tokens so that a single model can support both full-sequence and next-token generation \cite{chen2024diffusion}. Closest to our work, H{\o}eg et al.~\cite{hoeg_hybrid_2025} jointly generate a symbolic plan and continuous trajectory using discrete and continuous diffusion processes within a single transformer, with independent noise levels for the two modalities. On simulated and real tabletop tasks, the joint generation outperforms trajectory-only diffusion. Our work builds on this hybrid formulation, replacing diffusion with flow matching and studying the more data-limited setting of teleoperated mobile-manipulation demonstrations.

\paragraph*{Augmentation and replanning in imitation learning}
Relabeling and symmetry provide complementary ways to reuse demonstrations without collecting additional data. Hindsight experience replay relabels trajectories according to achieved goals \cite{andrychowicz2017hindsight}, while equivariant policies incorporate scene symmetries into the policy architecture \cite{wang2024equivariant}. Receding-horizon execution similarly improves robustness by repeatedly replanning over fixed-length action chunks \cite{zhao2023learning,chi2024diffusionpolicyvisuomotorpolicy}. This strategy has also been applied to real mobile manipulation using relatively small per-task demonstration sets supplemented by co-training data \cite{fu2024mobile}. Our work exploits object interchangeability for symbolic relabeling and performs replanning at symbolic action boundaries rather than at a fixed temporal horizon.

\section{Problem Formulation}
\label{sec:problem}

\begin{figure}[t]
    \centering
    \includegraphics[width=\linewidth]{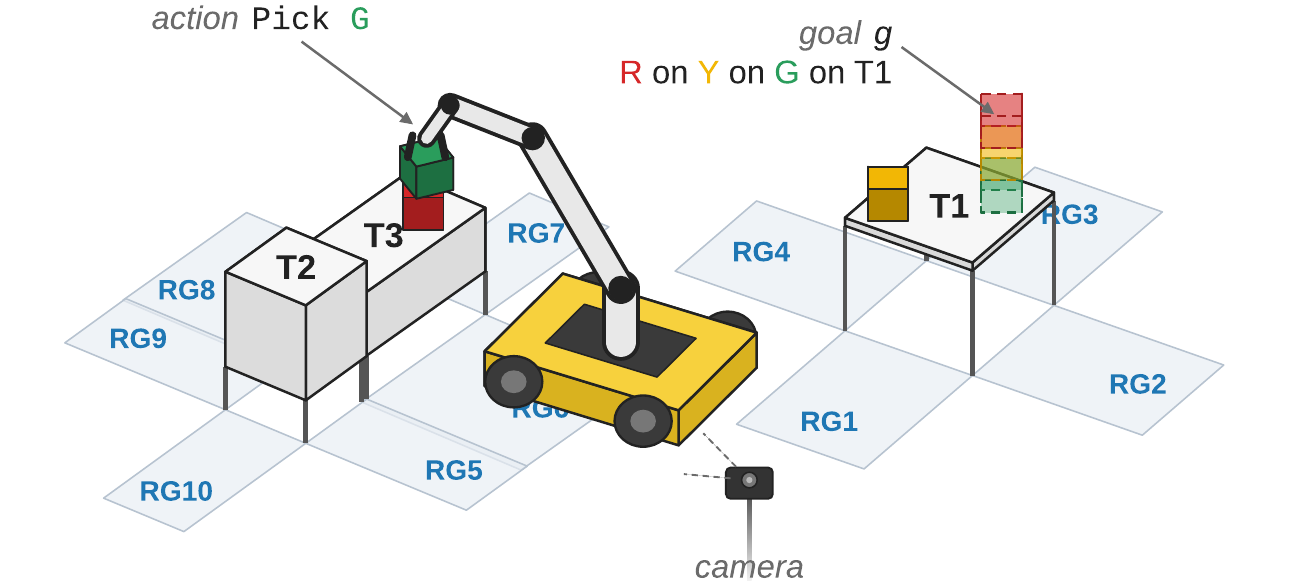}
    \caption{The running example. Three cubes R, Y, and G are placed on tables T1--T3, with approach regions RG1--RG10 located along the table sides. The observation $o$ contains the poses of the mobile manipulator and the cubes. A plan consists of a sequence of symbolic actions, such as \texttt{Pick G}, while the goal $g$ specifies the desired final configuration: R on Y on G on T1, illustrated by the ghost cubes.}
    \label{fig:scene}
\end{figure}
 
A task is given by the initial configuration, i.e., an observation $o$ of the scene, with the poses of the robot and of every movable object, and a symbolic goal $g$, i.e., a set of relations that must hold at the end.
In the running example (Fig.~\ref{fig:scene}), three cubes R, Y, and G can lie on three tables, and the goal is a tower in a given order on a given table.

A solution has two aligned levels, a discrete plan of $M$ actions and a continuous trajectory,
\begin{equation}
    \sigma = (a_{1}, \ldots, a_{M}), \qquad x_c = (x_{c,1}, \ldots, x_{c,M}),
    \label{eq:plan}
\end{equation}
where $x_{c,k}$ is the segment of waypoints that realizes action $a_k$, and $o_k$ denotes the observation at its start, with $o_1 = o$. Each symbolic action consists of an action type and its arguments, drawn from a finite set of entities, such as the objects and locations of the scene. In the running example, \texttt{Move\_to} takes an approach region, \texttt{Pick} a cube, and \texttt{Place} the currently held cube and a support, either a cube or the table on which to put it.
A waypoint is the commanded robot state, including the base pose, the end-effector pose, and the gripper command.

For training, each action is tokenized as its action type followed by one token per argument, and the tokens of all actions are concatenated and closed with an end-of-sequence token \texttt{EOS}, the end token.
The plan then becomes a token sequence $x_d = (\tau_1, \ldots, \tau_L)$ over a vocabulary $\mathcal{V}$ of action types, entities, and \texttt{EOS}.
A grammar $\mathcal{G}$ gives, for every prefix, the set $\mathcal{A}_\mathcal{G}(\tau_{1:i-1}) \subseteq \mathcal{V}$ of tokens allowed at position $i$: an action type or \texttt{EOS} if the preceding token completed an action, i.e. encoded its last argument, and otherwise an argument permitted for the action type currently being filled.
A sequence is \emph{grammatical} if
\begin{equation}
    \tau_i \in \mathcal{A}_\mathcal{G}(\tau_{1:i-1}) \ \ \forall i, \qquad \tau_L = \texttt{EOS} .
    \label{eq:grammatical}
\end{equation}

For evaluation, we use a symbolic model of the task to check whether a generated plan is valid. An abstraction $\phi$ maps an observation to a symbolic state; in the example that state contains the support of each cube, the cube held by the gripper, and the approach region of the robot. A transition $f$ maps a symbolic state and an applicable action to the next state. A plan is \emph{valid} for $(o, g)$ if
\begin{equation}
    s_1 = \phi(o), \qquad s_{k+1} = f(s_k, a_k), \qquad s_{M+1} \models g,
    \label{eq:valid}
\end{equation}
two valid plans for the same task are \emph{equivalent}, and a plan is \emph{successful} if the execution of $x_c$ from $o$ ends in a scene $o'$ with $\phi(o') \models g$.

Given teleoperated demonstrations $\mathcal{D} = \{(o, g, \sigma, x_c)\}$, labeled with their actions and associated $x_c$ segments, we learn a generative model
\begin{equation}
    p_\theta(x_d, x_c \mid o, g)
    \label{eq:model}
\end{equation}
whose samples for a new task are grammatical, valid, and successful, queried once per task or after every executed action (Section~\ref{sec:augmentation}). We assume that the poses are observed, that $\mathcal{V}$ and $\mathcal{G}$ are given, and that objects are rigid and picked one at a time. The symbolic model $(\phi, f)$ is used only for evaluation.

\section{Hybrid Flow Matching}
\label{sec:method}

\begin{figure*}[t]
    \centering
    \includegraphics[width=0.9\linewidth]{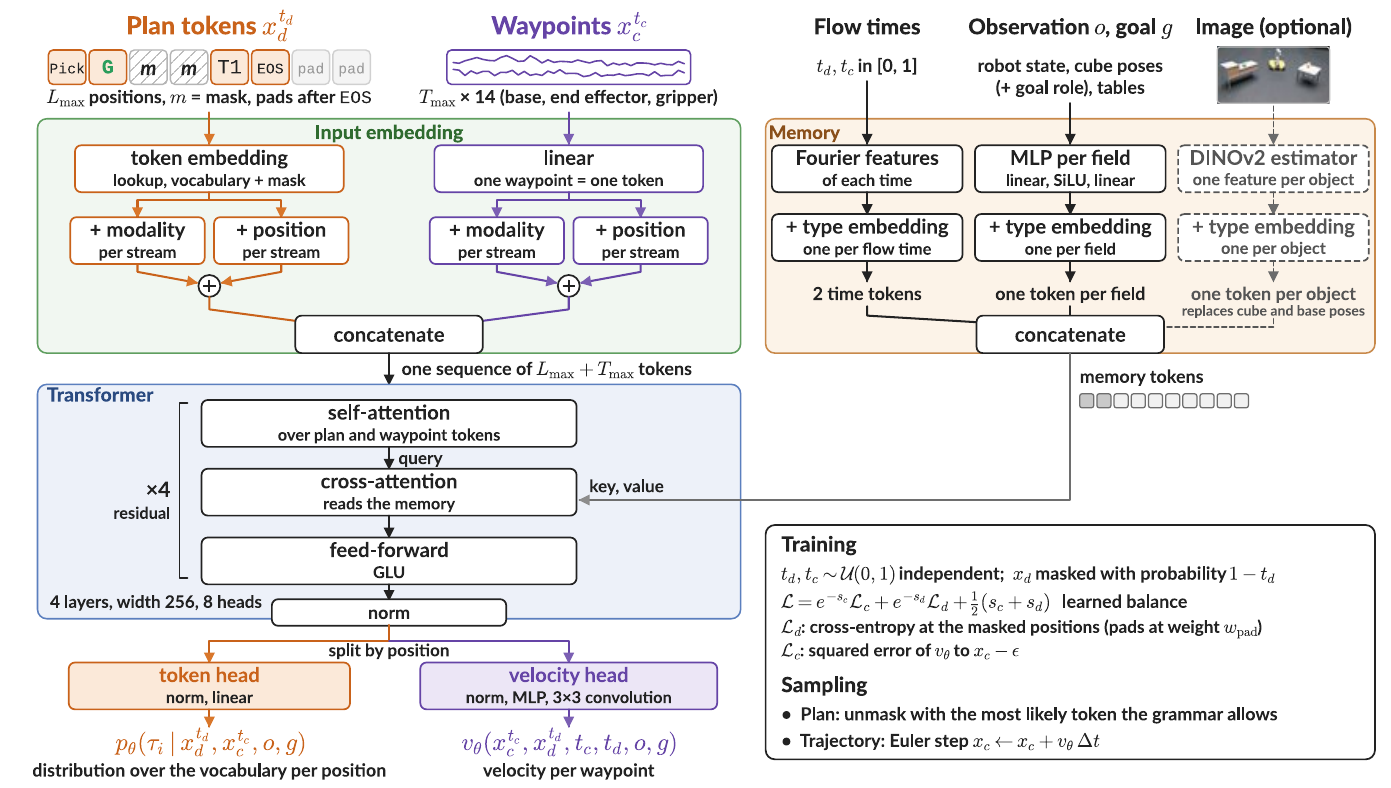}
    \caption{The network. The symbolic plan tokens and the waypoints get a token embedding or a linear layer, a learned modality and positional embedding, and form one sequence for the transformer. Its cross-attention reads the memory: one token per flow time (Fourier features) and one per observation component (an MLP), each with a type embedding. The token head gives $p_\theta$ at every symbolic position and the velocity head $v_\theta$ at every waypoint.}
    \label{fig:arch}
\end{figure*}

Sampling a plan from \eqref{eq:model} generates two data modalities, a discrete and a continuous one. 
The symbolic plan is generated with masked discrete flow matching, from a fully masked sequence. The trajectory is generated with continuous flow matching, from noise. Both flows run in parallel through the same network (Fig.~\ref{fig:arch}), so each modality conditions the other while it is generated. Each flow has its own time, $t_d$ and $t_c$, from $0$ (noise) to $1$ (data).

\subsection{Symbolic plan with discrete flow matching}
\label{sec:discrete}

The symbolic plan occupies the first $L$ positions of a sequence of fixed length $L_{\max}$, with the remaining positions filled by a pad token. Following \cite{gat2024discrete,campbell2024generative}, the noisy plan at time $t_d$ keeps each token with probability $t_d$ and replaces it by the mask token $m$ otherwise,
\begin{equation}
    \tau_i^{t_d} =
    \begin{cases}
        \tau_i & \text{with probability } t_d, \\
        m      & \text{otherwise},
    \end{cases}
    \qquad i = 1, \ldots, L_{\max}.
    \label{eq:mask}
\end{equation}
The network predicts a distribution $p_\theta(\tau_i \mid x_d^{t_d}, x_c^{t_c}, o, g)$ at every symbolic position, trained with the cross-entropy at the masked positions,
\begin{equation}
    \mathcal{L}_d = \frac{\sum_i w_i \, \mathbb{1}[\tau_i^{t_d} = m] \, \bigl(-\log p_\theta(\tau_i \mid x_d^{t_d}, x_c^{t_c}, o, g)\bigr)}{\sum_i w_i \, \mathbb{1}[\tau_i^{t_d} = m]},
    \label{eq:loss_d}
\end{equation}
where $w_i = 1$ for plan tokens and $w_i = w_{\mathrm{pad}}$ for pad position. Padding tokens are masked and supervised together with the plan, as in fixed-length masked models \cite{chang2022maskgit,sahoo2024simple}. This is what allows plans of different lengths. Without padding supervision, the sampler would reveal arbitrary tokens after \texttt{EOS}, and the plan positions that remain masked would see a context never encountered during training.

\paragraph*{Sampling}
Generation runs over $K$ steps on a time grid $0 = t^0 < \cdots < t^K = 1$ from a fully masked sequence. At each step, every masked position is revealed with probability $\Delta t / (1 - t)$ \cite{gat2024discrete} to its most likely token. We add three rules.

\emph{Grammar-constrained reveal.} When masks precede a position, the tokens allowed there depend on how the masks are filled.
We therefore compute the set $\mathcal{A}_\mathcal{G}(x_d^{t}, i)$ of tokens allowed at position $i$ under every completion of its prefix, and the reveal is
\begin{equation}
    \tau_i \leftarrow \arg\max_{\tau \in \mathcal{A}_\mathcal{G}(x_d^{t}, i)} p_\theta(\tau \mid x_d^{t_d}, x_c^{t_c}, o, g).
    \label{eq:grammar_reveal}
\end{equation}
If a revealed token becomes invalid after its left context changes, it is re-masked and predicted again. Unlike autoregressive grammar-constrained decoding \cite{scholak2021picard,geng2023grammar}, this operates on a partially masked sequence.

\emph{Freeze after the end token.} Once an \texttt{EOS} is visible, no position after it is revealed. These positions stay masked, and become pads at the end.

\emph{Remasking.} At three predefined steps, a fixed fraction of the least-confident visible tokens is re-masked and predicted again. This acts as a correction step \cite{campbell2022continuous,chang2022maskgit,wang2025remasking} and is used for all policies in our experiments.

\subsection{Motion with continuous flow matching}
\label{sec:continuous}

The trajectory has a fixed length of $T_{\max}$ waypoints, padded by repeating the last waypoint. With the linear path of flow matching \cite{flow-matching-lipman},
\begin{equation}
    x_c^{t_c} = (1 - t_c)\, \epsilon + t_c\, x_c, \qquad \epsilon \sim \mathcal{N}(0, I),
    \label{eq:path}
\end{equation}
the target velocity is $u = x_c - \epsilon$. The network predicts $v_\theta(x_c^{t_c}, x_d^{t_d}, t_c, t_d, o, g)$, and is trained with mean squared error over the set $\mathcal{T}$ of unpadded waypoints of dimension $n$,
\begin{equation}
    \mathcal{L}_c = \frac{1}{|\mathcal{T}| \, n} \sum_{t \in \mathcal{T}} \bigl\| v_{\theta, t} - u_t \bigr\|^2 .
    \label{eq:loss_c}
\end{equation}
At inference, the trajectory is initialized from Gaussian noise and integrated with Euler steps $x_c \leftarrow x_c + v_\theta \, \Delta t$, using the same sampling time grid as the symbolic plan.

\subsection{One network, one loss}
\label{sec:joint}

\paragraph*{Architecture}
We use a transformer \cite{peebles2023scalable,hoeg_hybrid_2025} with cross-attention (Fig.~\ref{fig:arch}). Each waypoint of $x_c^{t_c}$ and every position of $x_d^{t_d}$ forms one input token, augmented with learned positional and modality embeddings. The two sequences are concatenated, allowing self-attention to connect the modalities. Cross-attention conditions these tokens on a memory containing one token per component of the observation (the robot state, each object pose, and the goal) and one token per flow time.

\paragraph*{Flow times}
During training, $t_d, t_c \sim \mathcal{U}(0, 1)$ independently. With probability $1/4$ both are instead sampled from $\mathcal{U}(0, 0.2)$ \cite{hoeg_hybrid_2025}. Independent times teach the model to predict either modality at different generation stages of the other \cite{chen2024diffusion}, while the shared early range emphasizes the beginning of generation.

\paragraph*{Loss balance}
H{\o}eg et al.~\cite{hoeg_hybrid_2025} combine (\ref{eq:loss_d}) and (\ref{eq:loss_c}) with a fixed weight. The two terms differ in scale and evolve differently: the velocity error keeps decreasing while the token cross-entropy flattens, so a fixed weight leaves the plan under-weighted late in training. We learn their balance using uncertainty weighting \cite{kendall2018multi},
\begin{equation}
    \mathcal{L} = e^{-s_c} \mathcal{L}_c + e^{-s_d} \mathcal{L}_d + \tfrac{1}{2}(s_c + s_d),
    \label{eq:loss}
\end{equation}
where $s_c$ and $s_d$ are learned log-variance parameters.

\paragraph*{Joint sampling}
Each network evaluation produces both the symbolic token distributions and the continuous velocity. The symbolic plan takes its reveal step and the trajectory its Euler step from the same evaluation. During remasking, only $t_d$ is shifted for that evaluation, consistent with the independently sampled flow times used during training.

\subsection{Data augmentation}
\label{sec:augmentation}

Task and motion plans grow combinatorially with the number of objects, regions, and actions, while a demonstration covers one plan in one scene. We exploit two structures of the symbolic plan to generate additional training samples without collecting new demonstrations.

\begin{figure}[t]
    \centering
    \includegraphics[width=\linewidth]{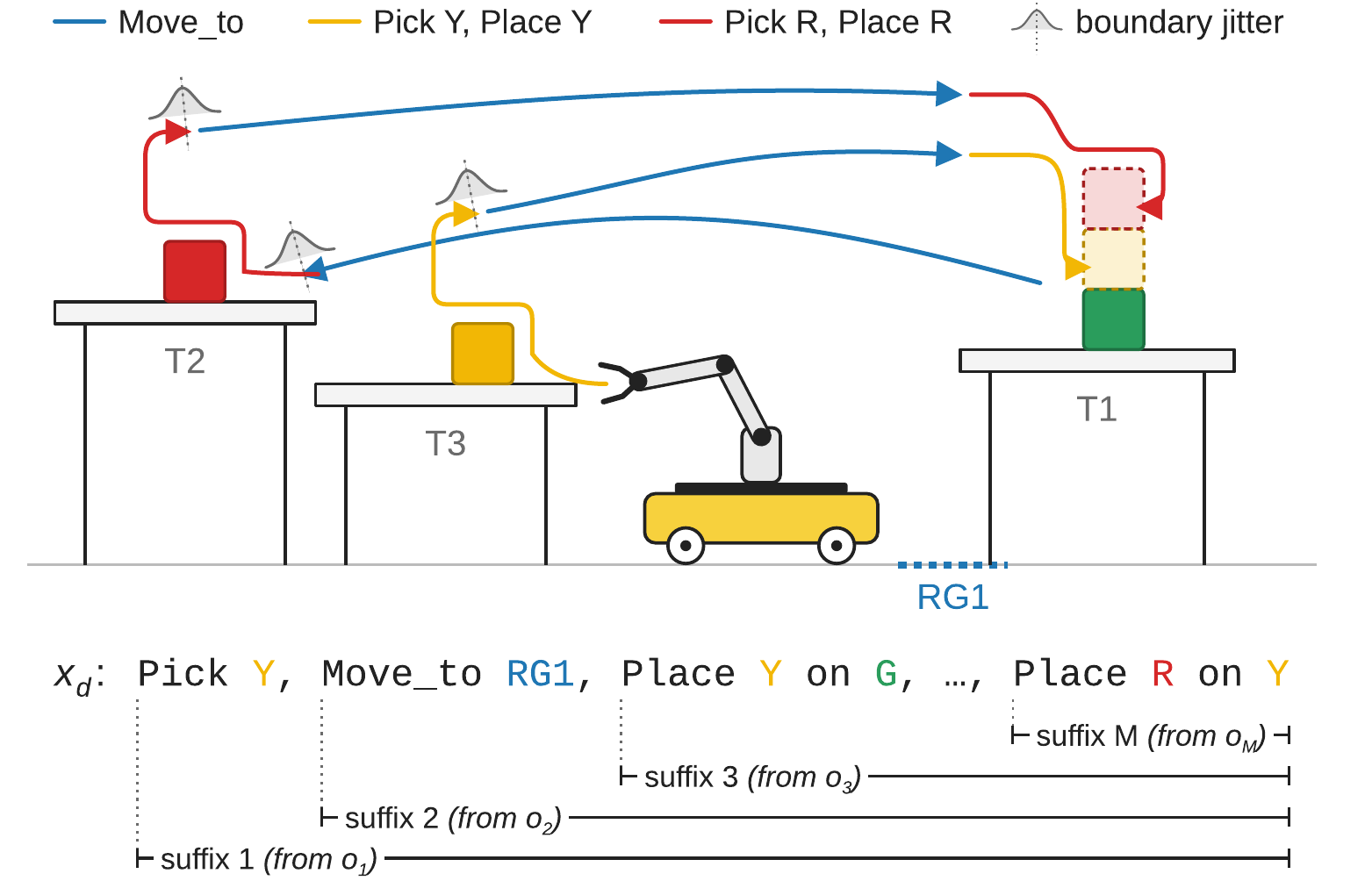}
    \caption{Plan suffixes of one demonstration. Top: the end-effector path over the scene, colored per action (blue \texttt{Move\_to}, yellow \texttt{Pick Y} and \texttt{Place Y}, red \texttt{Pick R} and \texttt{Place R}); the grey curves mark the boundary jitter around the action boundaries. Bottom: the plan $\sigma$ and the suffixes that start at the observations $o_1, \ldots, o_M$; each suffix is a training sample.}
    \label{fig:suffixes}
\end{figure}

\paragraph*{Plan suffixes}
The remainder of a demonstration after any action is itself a valid demonstration from the corresponding intermediate observation $o_k$. We therefore train on every suffix of every demonstration $(o, g, \sigma, x_c) \in \mathcal{D}$,
\begin{equation}
    \mathcal{D}^{+} = \bigl\{ (o_k, g, a_{k:M}, x_{c,k:M}) \bigr\}_{k=1}^{M},
    \label{eq:suffix}
\end{equation}
thus a demonstration of $M$ actions yields $M$ training samples (Fig.~\ref{fig:suffixes}). These samples support closed-loop execution at the symbolic-action level: the robot generates a plan and trajectory, executes only the segment corresponding to the first action, observes the resulting scene, and replans from that state. Consequently, execution errors such as failed grasps or placements can be handled by subsequent replanning rather than propagated through the remainder of an open-loop trajectory.

Constructing the suffixes requires the action boundaries of each demonstration, which are available from the recorded labels. To reduce sensitivity to the exact boundary location, we can use \emph{boundary jitter}, shifting the start of a suffix by a small number of waypoints while keeping the same symbolic plan. This exposes the model to slightly different observations around each action transition.

\begin{figure}[t]
    \centering
    \includegraphics[width=\linewidth]{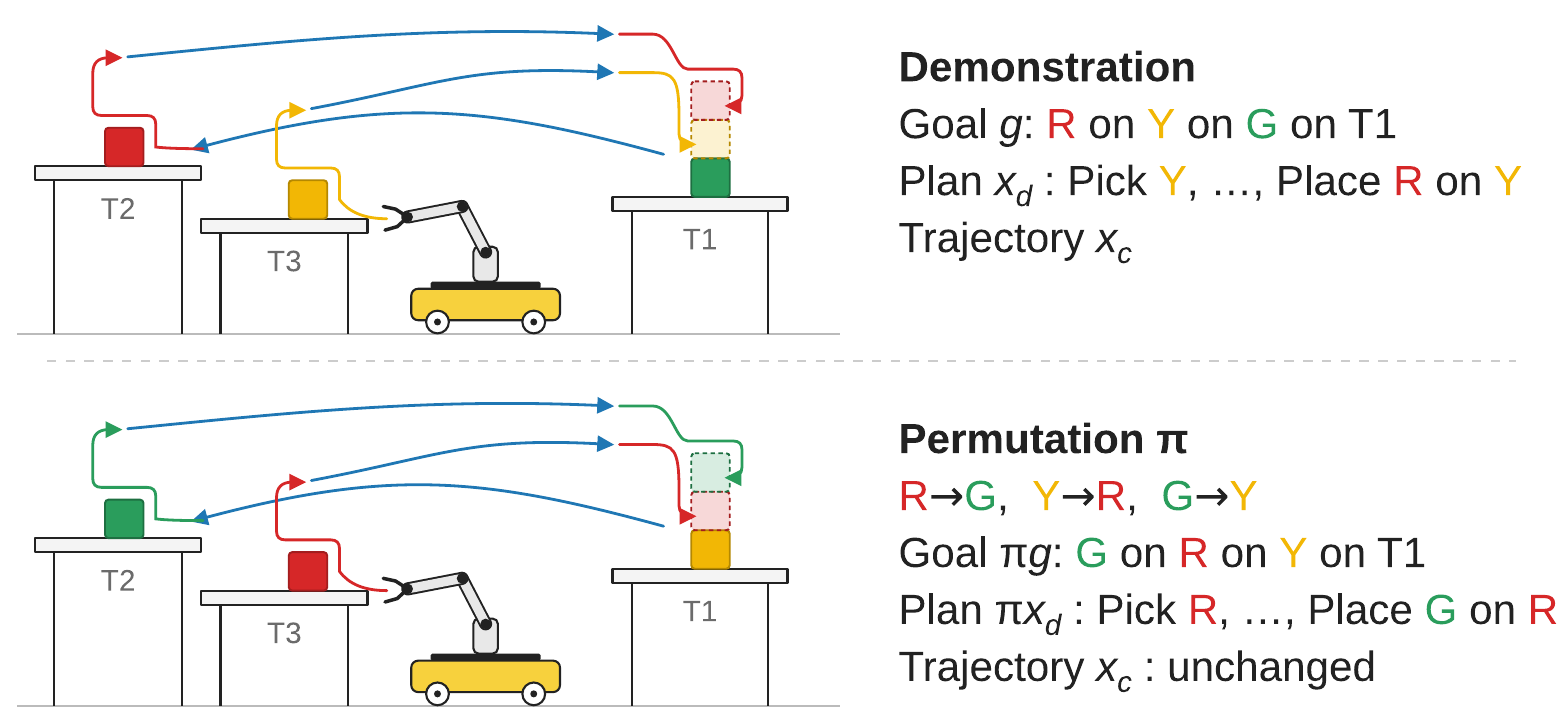}
    \caption{Symbolic permutation of the same demonstration. Renaming the cubes (R to G, Y to R, G to Y) in the goal, the plan, and the observation gives a demonstration of another tower order with the same trajectory.}
    \label{fig:permutations}
\end{figure}

\paragraph*{Symbolic permutation}
Entities of the same kind are interchangeable, such as the cubes of the running example. For a permutation $\pi$ of interchangeable entities, consistently renaming their representation in the observation, goal, and symbolic plan yields
\begin{equation}
    (o, g, \sigma, x_c) \ \mapsto \ (\pi o, \ \pi g, \ \pi \sigma, \ x_c),
    \label{eq:perm}
\end{equation}
while the continuous trajectory remains unchanged because the renamed entities occupy the same physical poses. Training on these permutations exposes the planner to every goal assignment of the entities in each demonstration (Fig.~\ref{fig:permutations}); the object permutations of Section~\ref{sec:intro} are this augmentation applied to the cubes. This is related to goal relabeling \cite{andrychowicz2017hindsight} and provides a data-level analogue of equivariant modeling \cite{wang2024equivariant}. The augmentation requires an observation in which the permuted entities have separate representations. The goal is encoded per entity, together with its pose, so that each entity's task role is associated with its spatial state.

\section{Experiments}
\label{sec:results}

The experiments answer three questions. Does the planner produce valid plans and successful executions on scenes it has not seen, and how does it compare with the two closest alternatives, motion-only flow matching \cite{black2024pi0, braun2024riemannian} and hybrid diffusion \cite{hoeg_hybrid_2025}? What does each part of the training and generation scheme contribute? Does the planner run on the real robot?

\subsection{Setup}
\label{sec:setup}

\begin{figure}[t]
    \centering
    \includegraphics[width=\linewidth]{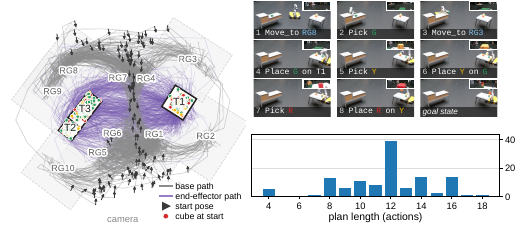}
    \caption{The 109 (+12 validation) demonstrations. Left: top view with the tables T1--T3 and the approach regions RG1--RG10: base paths (grey), end-effector paths (violet), start poses (arrows), and initial cube positions (dots). Right: one demonstration represented by observations \(o_1,\ldots,o_8\) at the action boundaries, together with the corresponding actions, wrist-camera views, and the distribution of plan lengths at the bottom right.}
    \label{fig:dataset}
\end{figure}

\paragraph*{Task and robot}
The robot is a holonomic mobile base (Clearpath Dingo-O) with a six-joint arm (Kinova Gen3 Lite); the cubes have a side of 6.5 cm and the tables differ in height and size (Fig.~\ref{fig:scene}). The vocabulary has three action types, the three cubes, the three tables, ten approach regions, and the end token \texttt{EOS}. A waypoint is the robot's base pose with respect to the floor plane, the end-effector pose in 3D, and the gripper state, 14 numbers.

\paragraph*{Demonstrations}
109 demonstrations were recorded using teleoperation, with the cube and robot poses from a motion capture system, and 12 additional demonstrations were recorded and held out for validation. The cubes start on any table, alone or stacked, and the robot anywhere in the room; T1 starts empty, with G on it, or with a ``wrong'' cube on it (Fig.~\ref{fig:dataset}). Demonstrations contain 4--18 actions and 17--69 waypoints, sampled approximately every 0.8 s, with each action labeled by its starting waypoint. %
\paragraph*{Training and inference}
The transformer has four layers of width 256 with eight heads, 7.5 million parameters. It is trained for 20\,000 steps with batches of 45 samples and $w_{\mathrm{pad}} = 0.1$, with the weights averaged over training, in about 20 minutes on one NVIDIA RTX 3090. Inference uses $K = 100$ steps. Every configuration is trained with three seeds; we report the mean and the standard deviation over them.

\begin{figure}[t]
    \centering
    \includegraphics[width=\linewidth]{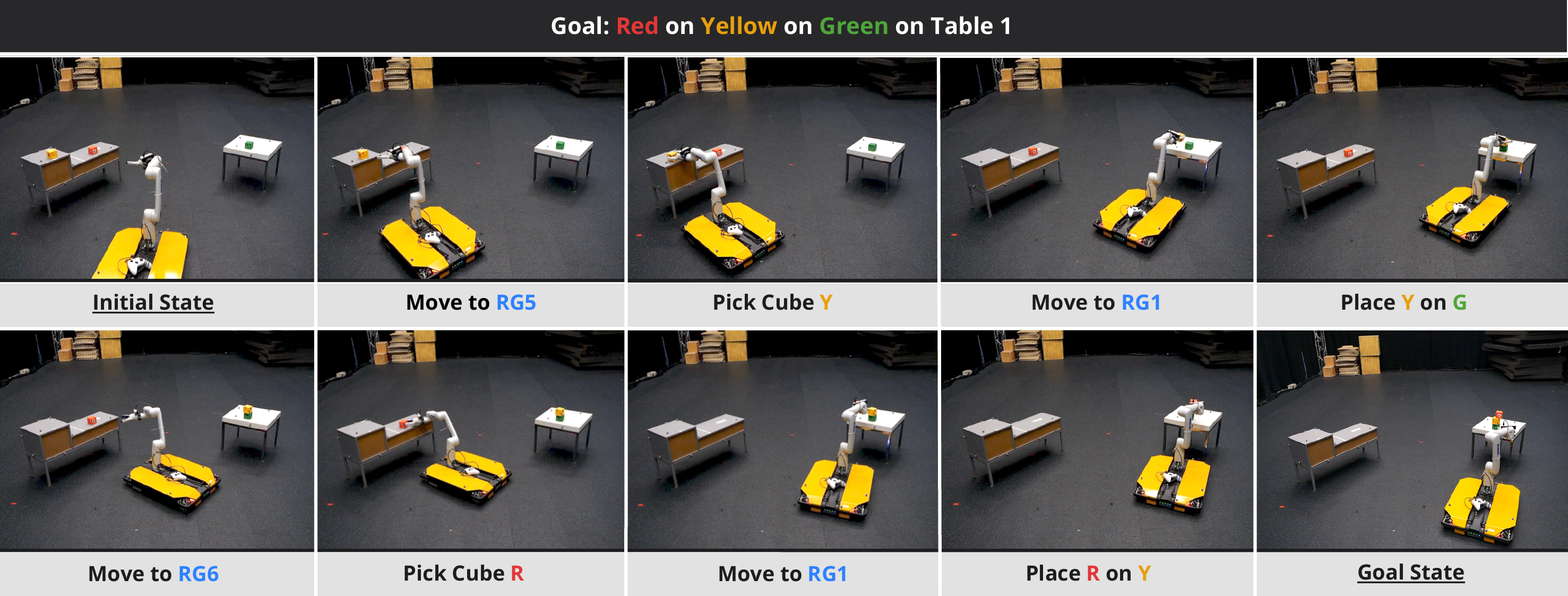}
    \caption{A run on the real robot with the goal R on Y on G on table T1 (cubes R red, Y yellow, G green). After every action the policy observes the new scene and samples a new plan and trajectory. Here every action succeeds, so the plan stays the same from one planning step to the next and the run looks like an open-loop execution.}
    \label{fig:real_run}
\end{figure}

\paragraph*{Metrics}
A sampled plan is \emph{valid} under the symbolic model of Section~\ref{sec:problem}, evaluated from the scene at the start of the task. Execution is evaluated in a kinematic simulation of the scene: the sampled waypoints move the robot, a cube within 9 cm of the end effector attaches when the gripper closes, and a released cube settles on the support beneath it. Tolerances are set to the 99th percentile of the demonstrations. The \emph{task completion} rate is the fraction of samples after which the goal tower stands, whatever plan was executed. In the \emph{open loop}, the whole sampled trajectory is executed; in \emph{closed loop}, only the first action segment is executed before replanning, with a budget of three extra actions. Results are reported overall and by the number of actions remaining to the goal, $d$.

\paragraph*{Policies}
\emph{Ours} is the hybrid flow matching planner of Section~\ref{sec:method}, trained on the suffix samples with the full training and generation scheme. \emph{Ours (base)} has a fixed loss weight and pad tokens without supervision; it uses grammar-constrained generation and the freeze after the end token, so that the comparison isolates the training changes. \emph{Ours + permutations} is the goal-conditioned policy trained on the six permutations of the three cubes; it builds any tower order and is scored on the recorded one only. \emph{Motion-only FM} (flow matching) is continuous flow matching on the waypoints alone \cite{black2024pi0, braun2024riemannian}, the same network without the plan tokens, with a training sample starting at every waypoint of the demonstrations, not only at the action boundaries; without a plan, its execution follows its own gripper commands (closing picks the nearest cube, opening releases it). \emph{Hybrid diffusion} is the formulation of H{\o}eg et al.~\cite{hoeg_hybrid_2025} on our network and data (diffusion on the waypoints, masked diffusion on the tokens, their loss weight, sampler, and optimizer), trained as published on whole demonstrations or on our suffix samples.

\subsection{Comparison}
\label{sec:comparison}

\begin{figure*}[t]
    \centering
    \includegraphics[width=\linewidth]{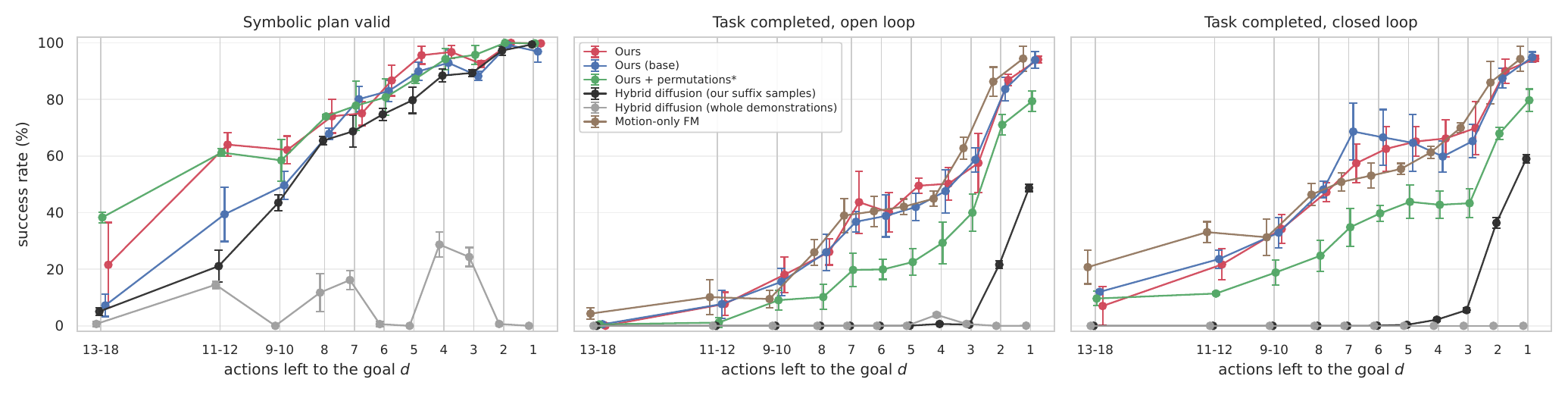}
    \caption{Success against the number of actions left to the goal, on the 12 held-out demonstrations (mean and standard deviation over three training seeds, 50 samples per task). Values of $d$ with few demonstrations are grouped: 13--18, 11--12, and 9--10 hold 850, 750, and 900 tasks, the other values 550 to 600 each. Left: the sampled plan is valid. Middle: the tower stands after the whole sampled trajectory is executed. Right: the tower stands when the policy replans after every action. \emph{Hybrid diffusion} is the formulation of H{\o}eg et al.~\cite{hoeg_hybrid_2025}, trained on our suffix samples or on whole demonstrations. \emph{Motion-only FM} is flow matching on the trajectory alone; it has no plan and no left panel. *The goal-conditioned policy builds any of the six tower orders and is scored on the recorded one.}
    \label{fig:per_d}
\end{figure*}

\begin{table}[t]
    \caption{Performance in percent on the 12 held-out demonstrations, averaged over 3 training seeds and 50 samples per task.}
    \label{tab:results}
    \centering
    \setlength{\tabcolsep}{3pt}
    \resizebox{\columnwidth}{!}{%
    \begin{tabular}{@{}lcccc@{}}
        \hline
        & \multicolumn{2}{c}{Symbolic plan valid$^\dagger$ (\%)} & \multicolumn{2}{c@{}}{Task completed$^\ddagger$ (\%)} \\
        Policy & all $d$ & $d \geq 13$ & open loop & closed loop \\
        \hline
        Motion-only FM & -- & -- & 39 $\pm$ 2 & 52 $\pm$ 2 \\
        Hybrid diffusion \cite{hoeg_hybrid_2025}: & & & & \\
        \quad whole demonstrations & \phantom{0}8 $\pm$ 1 & \phantom{0}1 & \phantom{0}0 & \phantom{0}0 \\
        \quad our suffix samples & 62 $\pm$ 1 & \phantom{0}5 & \phantom{0}6 $\pm$ 0 & \phantom{0}9 $\pm$ 0 \\
        Ours & 76 $\pm$ 1 & 33 & 40 $\pm$ 3 & 53 $\pm$ 3 \\
        Ours + permutations* & 76 $\pm$ 1 & 53 & 26 $\pm$ 3 & 36 $\pm$ 3 \\
        \hline
        \multicolumn{5}{@{}l}{\emph{The scheme one step at a time, from \emph{Ours (base)}}} \\
        \hline
        Unconstrained generation & 55 $\pm$ 1 & \phantom{0}5 & 30 $\pm$ 2 & 41 $\pm$ 4 \\
        + grammar and freeze (base) & 68 $\pm$ 2 & 10 & 38 $\pm$ 3 & 53 $\pm$ 5 \\
        + pad supervision & 71 $\pm$ 4 & 20 & 38 $\pm$ 3 & 53 $\pm$ 3 \\
        + learned loss balance = Ours & 76 $\pm$ 1 & 33 & 40 $\pm$ 3 & 53 $\pm$ 3 \\
        \hline
    \end{tabular}}
    \par\vspace{2pt}
    \begin{minipage}{\columnwidth}\scriptsize
        $^\dagger$The sampled plan is valid under the symbolic model, over all tasks, or over those with $d \geq 13$ actions to the goal. \par
        $^\ddagger$The tower stands after execution in the kinematic simulation, of the whole sampled trajectory (open loop) or with replanning after every action (closed loop). \par
        *Goal-conditioned: trained on all tower order permutations (Fig.~\ref{fig:permutations}), scored on the recorded one.
    \end{minipage}
\end{table}

\paragraph*{Plans}
\emph{Ours} produces valid symbolic plans in about three quarters of the samples (Table~\ref{tab:results}). Validity is near perfect with four or fewer actions remaining, but falls to about one third beyond 13 actions (Fig.~\ref{fig:per_d}, left). On training demonstrations of similar length, validity remains above 95\% up to 17 actions, indicating that long-horizon failures arise mainly from generalization to unseen scenes rather than model capacity. \emph{Hybrid diffusion} trained on our suffix samples plans, at the symbolic level, nearly as well as \emph{Ours (base)}, whereas training on whole demonstrations performs poorly on tasks that start mid-way through a demonstration, since they are absent from its training distribution. \emph{Ours + permutations} preserves symbolic plan validity and has the best long-horizon validity, while building any of the six tower orders.

\paragraph*{Execution}
\emph{Ours} and \emph{Ours (base)} achieve similar task completion rates. Most execution failures result from grasp or placement errors of a few centimeters, which improved symbolic validity alone cannot correct. \emph{Motion-only flow matching} performs similarly, showing that generating the plan jointly adds an explicit plan, action-level replanning, and goal conditioning rather than motion precision. \emph{Hybrid diffusion} shows larger trajectory error, with a median end-effector error of 0.71 m compared with 0.25 m for ours. %

\paragraph*{Closed loop}
Replanning after every action raises the task completion rate by 13\%  overall and from 0 to 10--30\% at long horizons (Fig.~\ref{fig:per_d}, right). The resulting rate matches the rate predicted from the per-action success rates on demonstrated scenes, so the states reached during replanning are as usable as the demonstrated ones. Moreover, 8--10\% of rollouts recover after a failed action. The goal-conditioned policy shows a similar improvement.

\subsection{Ablations}
\label{sec:ablations}

The lower part of Table~\ref{tab:results} adds each component sequentially. Grammar-constrained generation and freezing after \texttt{EOS} raise plan validity from 55\% to 68\% and the closed-loop task completion rate from 41\% to 53\%, because a rollout that ends in an ungrammatical plan cannot continue. Pad supervision further improves validity by 3\%, particularly at long horizons, while learned loss balancing raises overall validity to 76\% and long-horizon validity to 33\%. The open-loop task completion rate changes little, indicating that these components primarily improve symbolic planning rather than motion accuracy.

The goal-conditioned policy keeps plan validity of \emph{Ours} and achieves the best long-horizon validity of the study, but its task completion rate drops by 14\%. This decrease is attributable to the object-wise observation encoding required for permutation augmentation: applying the same encoding to \emph{Ours} produces a comparable loss in execution accuracy.

\subsection{Real robot}
\label{sec:real}

\begin{figure}[t]
    \centering
    \includegraphics[width=\linewidth]{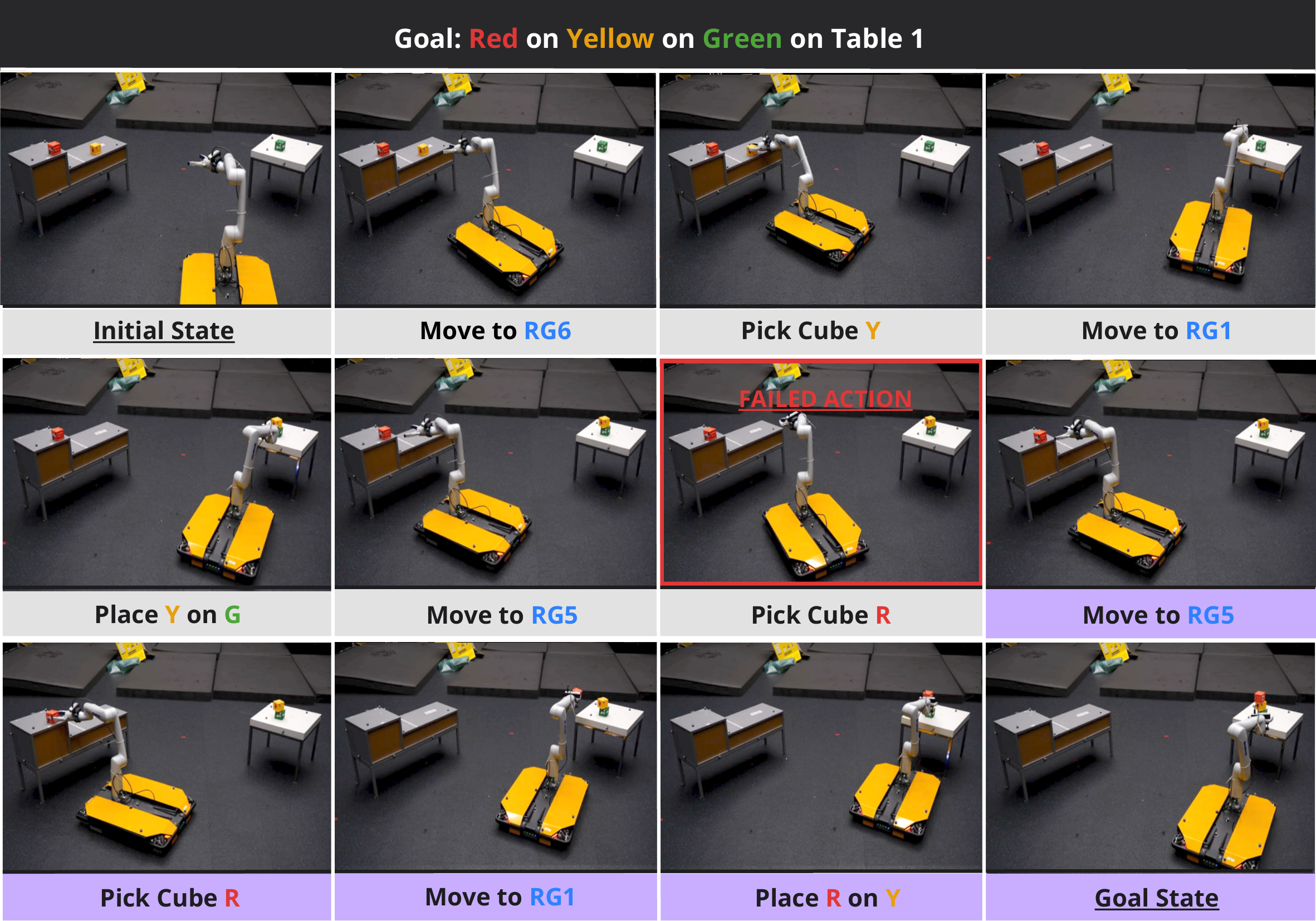}
    \caption{The same setup with a failure. The grasp of R misses; at the next planning step, the policy sees R still on the table, plans the pick again, and finishes the tower. Replanning after every action turns a missed grasp into a retry instead of a failed run.}
    \label{fig:real_run_replan}
\end{figure}

The policy runs on the robot from the motion capture poses, with the end-effector waypoints converted to joint targets by inverse kinematics and the base waypoints sent to the base controller, and it replans after every action. Figure~\ref{fig:real_run} shows a run in which every action succeeds, and Fig.~\ref{fig:real_run_replan} a run in which a missed grasp is retried. The robot was tested in 30 unseen cube placements: the tower was built in about 30\% of the runs from short plans ($d\leq 6$) and in 10\% from long ones ($d\geq 8$). The failures are the same as in the kinematic simulation, grasps missed by a few centimeters and placements off the support.

\paragraph*{Perception}
The results above take the cube poses from motion capture. To check that the policy does not depend on them and works from image perception, we trained it on image features instead: the object pose tokens are replaced by one token per object from a DINOv2 keypoint estimator \cite{oquab2023dinov2}, trained on images from a fixed camera recorded alongside the demonstrations, with the same held-out split.
The policy keeps its plan validity (78\% against 76\%) and loses 6\% of open-loop task completion rate.

\section{Conclusions}
\label{sec:conclusion}

We presented a hybrid flow matching planner, trained on 109 teleoperated demonstrations of a mobile manipulator, that jointly generates symbolic plans and continuous trajectories. Plan-suffix and object-permutation augmentations improve data efficiency, enabling action-level replanning and goal conditioning. The proposed training and generation scheme increases the plan validity on held-out scenes to 76\% while closed-loop replanning after every action raises the task completion rate from 40\% (open loop) to 53\% in a kinematic simulation, and allows recovery from failed actions on the real robot. Compared with the two closest alternatives, the method matches the task completion rate of \emph{motion-only flow matching} while providing an explicit plan, and improves trajectory accuracy over \emph{hybrid diffusion}.

\bibliographystyle{IEEEtran}
\bibliography{references}

\end{document}